\documentclass[conference]{IEEEtran}
\IEEEoverridecommandlockouts

\usepackage{url}
\usepackage{cite}
\usepackage{amsmath,amssymb,amsfonts}
\usepackage{algorithmic}
\usepackage{graphicx}
\usepackage{textcomp}
\usepackage{xcolor}
\usepackage{algorithm}
\usepackage{graphicx}
\usepackage{subfig}  %

\title{Sample Efficient Experience Replay in Non-stationary Environments}
\author{
\IEEEauthorblockN{
Tianyang Duan\IEEEauthorrefmark{1},
Zongyuan Zhang\IEEEauthorrefmark{1},
Songxiao Guo\IEEEauthorrefmark{1},
Yuanye Zhao\IEEEauthorrefmark{2},
Zheng Lin\IEEEauthorrefmark{3},
Zihan Fang\IEEEauthorrefmark{4},\\
Yi Liu\IEEEauthorrefmark{4},
Dianxin Luan\IEEEauthorrefmark{5},
Dong Huang\IEEEauthorrefmark{6},
Heming Cui\IEEEauthorrefmark{1},
Yong Cui\IEEEauthorrefmark{7}
}
\IEEEauthorblockA{\IEEEauthorrefmark{1}Department of Computer Science, The University of Hong Kong, Hong Kong, China}
\IEEEauthorblockA{\IEEEauthorrefmark{2}College of International Education, Hebei University of Economics and Business, China}
\IEEEauthorblockA{\IEEEauthorrefmark{3}Department of Electrical and Electronic Engineering, The University of Hong Kong, Hong Kong, China}
\IEEEauthorblockA{\IEEEauthorrefmark{4}Department of Computer Science, City University of Hong Kong, Hong Kong, China}
\IEEEauthorblockA{\IEEEauthorrefmark{5}Institute for Imaging, Data and Communications, University of Edinburgh, UK}
\IEEEauthorblockA{\IEEEauthorrefmark{6}School of Computing, National University of Singapore, Singapore}
\IEEEauthorblockA{\IEEEauthorrefmark{7}Department of Computer Science and Technology, Tsinghua University, China}
}

\begin{document}

\maketitle
\begin{abstract}
Reinforcement learning (RL) in non-stationary environments is challenging, as changing dynamics and rewards quickly make past experiences outdated. Traditional experience replay (ER) methods, especially those using TD-error prioritization, struggle to distinguish between changes caused by the agent's policy and those from the environment, resulting in inefficient learning under dynamic conditions. To address this challenge, we propose the Discrepancy of Environment Dynamics (DoE), a metric that isolates the effects of environment shifts on value functions. Building on this, we introduce Discrepancy of Environment Prioritized Experience Replay (DEER), an adaptive ER framework that prioritizes transitions based on both policy updates and environmental changes. DEER uses a binary classifier to detect environment changes and applies distinct prioritization strategies before and after each shift, enabling more sample-efficient learning. Experiments on four non-stationary benchmarks demonstrate that DEER further improves the performance of off-policy algorithms by 11.54\% compared to the best-performing state-of-the-art ER methods.
\end{abstract}

\section{Introduction}
\label{sec:intro}
\vspace{-0.5em}
Reinforcement Learning (RL)~\cite{zhang2025robust,duan2025rethinking} is a powerful dynamic sequential decision-making planning method widely applied in real-world scenarios \cite{tang2025deep}. However, practical environments are often non-stationary, with changing environment dynamics and reward signals over time \cite{khetarpal2022towards}. This makes adaptive planning essential to handle real-world complexity and unpredictability.

Off-policy reinforcement learning (RL) leverages historical experiences to address the challenges of sparse and costly sampling in high-dimensional, continuous action spaces \cite{yarats2021improving}.  This efficiency is primarily facilitated by Experience Replay (ER), which stores and reuses past transitions to break temporal correlations and enhance data efficiency. Recent research has improved ER by introducing non-uniform sampling, with temporal difference error (TD-error) based prioritization significantly enhancing sample efficiency in stationary environments \cite{sun2020attentive,li2024prioritized}. TD-error quantifies the disparity between the estimated return and the actual return., enabling the agent to identify and prioritize transitions that offer the most informative experiences. As a result, transitions with higher TD-error are sampled more frequently, accelerating convergence and improving performance.

However, in non-stationary environments, historical experiences quickly become outdated, which can undermine effective sample selection and mislead learning \cite{rahimi2023replay}. Most existing methods overlook the negative impact of obsolete transitions, especially as TD-error is influenced by both environmental changes and policy updates. After the value function adapts to a new environment, transitions collected before the change often show higher TD-errors, and prioritizing these can amplify irrelevant experiences, reducing training efficiency and performance. Similar issues arise with reward-based \cite{cao2019high} or frequency-based sampling \cite{sun2020attentive}. Ultimately, prioritization strategies that focus only on policy improvement fail to account for dynamic environments and cannot accurately assess the relevance of stored transitions.

To address the challenge of non-stationary environments, we propose the Discrepancy of Environment (DoE), a principled metric quantifying the impact of environmental changes on state transitions. By measuring the deviation between action-value functions before and after a dynamics shift—while excluding policy improvement effects—DoE precisely attributes value changes to underlying environment dynamics. Building on this, we propose Discrepancy of Environment Prioritized Experience Replay (DEER), a sample-efficient replay framework that adaptively prioritizes experience samples for both policy optimization and environment adaptation. DEER leverages a binary classifier to detect dynamic shifts by estimating reward sequence distributions across adjacent time windows, and applies distinct prioritization strategies to transitions before and after detected changes. Pre-change transitions with lower DoE are deemed more relevant and prioritized, while post-change transitions are ranked using a hybrid of TD-error and real-time DoE-based density differences. This approach maintains replay buffer diversity and dynamically allocates sampling priorities to meet agent demands. Extensive experiments on four non-stationary Mujoco benchmarks show that, compared with the state-of-the-art experience replay methods, DEER further improves the performance of off-policy algorithms by 11.54\%. Furthermore, under extreme non-stationary settings (200\% shift), DEER achieves an additional 22.53\% performance improvement for offline policy algorithms relative to the best ER method.
\vspace{-1em}
\section{Related Work}
\vspace{-1em}

Experience replay enhances sample efficiency and learning stability by storing and resampling past experiences \cite{lin1992self}.  While it offers the potential to accelerate adaptation in non-stationary environments, this advantage remains insufficiently explored. The predominant approach, PER, samples transitions according to their TD-error, yielding significant gains across diverse RL benchmarks \cite{sun2020attentive,li2024prioritized}. Beyond TD-error based sampling, several alternative strategies have been developed: PSER increases the priority of transitions preceding salient events \cite{brittain2019prioritized}; ReF-ER restricts sampling to “near-policy” transitions defined by policy similarity \cite{novati2019remember}; and AER emphasizes transitions that are similar to the agent’s current state \cite{sun2020attentive}. Additionally, HER and its variants \cite{andrychowicz2017hindsight,luo2023relay,sayar2024hindsight} address sparse-reward environments by relabeling goals, thereby providing richer reward signals. RB-PER \cite{li2021revisiting} assigns higher priority to infrequently sampled transitions, promoting sampling diversity and enhancing adaptation to non-stationarity.
\vspace{-2em}
\section{Methodology}
\vspace{-0.5em}
\subsection{Problem Formalization}
\vspace{-0.5em}
RL in non-stationary environments is formulated as a family of Markov decision processes (MDPs) \cite{padakandla2021survey}, denoted by $\left \{ \left \langle \mathcal{S},\mathcal{A} ,P_i,R_i,T_i,\gamma   \right \rangle \right \} _{i=0}^{\infty }$. Here, $\mathcal{S}$ is the state space, and $\mathcal{A}$ is the action space. At each time step $t$, the agent selects an action $a_t \in \mathcal{A}$ using a policy $\pi(a_t \mid s_t)$, transitions to the next state $s{t+1} \in \mathcal{S}$ according to the state-transition probability $P(s_{t+1} \mid s_t, a_t)$, and receives a reward $r_t = R(s_t, a_t)$. The state-transition function in non-stationary environments is defined as:
\begin{equation}
\small
\label{eq:1}
P\left(s_{t+1} \mid s_t, a_t\right) = 
\begin{cases}
\begin{aligned}
&P_0\left(s_{t+1} \mid s_t,a_t\right), \, 0\le t \le T_0 \\
&\dots \\
&P_i\left(s_{t+1} \mid s_t,a_t\right), \,  T_{i-1} < t \le T_i \\
&\dots 
\end{aligned}
\end{cases}
\end{equation}
where $T_i$ denote the time step at which the environment dynamics change. The reward function of non-stationary environments is defined as: 
\begin{equation}
\small
\label{eq:2}
R\left(s_t, a_t\right) = 
\begin{cases}
\begin{aligned}
&R_0\left( s_t,a_t\right), \, 0\le t \le T_0 \\
&\dots \\
&R_i\left( s_t,a_t\right), \,  T_{i-1} < t \le T_i \\
&\dots 
\end{aligned}
\end{cases}
\end{equation}

The objective of an agent is to learn a policy $\pi $ to maximize the expected discounted return $\mathbb{E}_\pi \left [  {\textstyle \sum_{k=0}^{\infty }\gamma ^kr_{t+k}} \right ] $, where $\gamma $ denotes the discount factor.

\subsection{Discrepancy of Environment (DoE)}
\label{sec:DoE}
\vspace{-0.5em}
An effective experience replay mechanism for non-stationary environments should prioritize transitions that help the agent quickly adapt to changes in dynamics. To achieve this, we track shifts in environmental dynamics by analyzing agent-environment trajectories, $\tau = [s_0, a_0, s_1, a_1, \ldots]$, generated under the current policy $\pi(a \mid s)$. Since the state-transition function captures the environment’s dynamics, changes in these trajectories provide reliable signals of dynamic shifts:
\begin{equation}
\small
\label{eq:the probability density of trajectory}
P\left ( \tau  \right ) =P\left ( s_0 \right ) \prod_{t=0}^{\infty } \pi \left ( a_t\mid s_t \right )P\left ( s_{t+1}\mid s_t,a_t \right ) ,
\end{equation}
where $P(s_0)$ denotes the initial state distribution. Given a fixed initial state distribution and policy, any changes in environment dynamics directly affect the trajectory distribution. Since rewards are determined solely by the reward function, the probability density of a reward sequence, $\mathbf{r}_\tau = [r_0, r_1, \dots]$, is induced by the trajectory distribution. Therefore, changes in environment dynamics can be approximated by analyzing variations in the reward sequence.


To this end, we propose an environmental dynamics monitoring method based on density-ratio estimation (DRE) \cite{hillman2021optimizing}, leveraging a binary classifier trained on reward sequences. Specifically, at each time step $t$, we construct two adjacent sliding windows: a reference window $\mathbf{\widetilde{r}}_{rf} =\left \{ \mathbf{r} _{i:j} \mid t-2m+1\le i< j\le t-m \right \}$ labeled with $l=0$, and a test window $\mathbf{\widetilde{r}} _{te} =\left \{ \mathbf{r} _{i:j}\mid t-m<  i< j\le t \right \}$ labeled with $l=1$, where $m$ denotes the window size. The binary classifier is trained to distinguish between samples from the reference and test windows, thereby estimating the density ratio between their distributions. Formally, the classifier models the conditional distribution as follows:
\begin{equation}
\small
\label{eq:12}
P\left ( \mathbf{r}\mid f\left ( \mathbf{r}  \right ) =l \right )  = 
\begin{cases}
  P_{rf}\left ( \mathbf{r} \right ) &  \text{ if } l=0 \\
  P_{te}\left ( \mathbf{r} \right ) &  \text{ if } l=1
\end{cases} 
\end{equation}
where $P_{rf}\left ( \cdot  \right ) $ and $P_{te}\left ( \cdot  \right ) $ denote the density functions of the reference and test windows, respectively, and $f\left ( \cdot  \right ) $ represents the binary classifier. We employ a multilayer perceptron (MLP) as a binary classifier and optimize its parameters by minimizing the following cross-entropy loss function:
\begin{equation}
\small
\label{eq:13}
\mathcal{L} \left ( f \right ) =-\frac{1}{\left | \mathbf{\widetilde{r}} _{rf} \right | } \sum_{\mathbf{r}\in \mathbf{\widetilde{r}} _{rf} }\log{\left ( 1-f\left ( \mathbf{r}  \right )  \right ) } -\frac{1}{\left | \mathbf{\widetilde{r}} _{te} \right | } \sum_{\mathbf{r}\in \mathbf{\widetilde{r}} _{te} }\log{ f\left ( \mathbf{r}  \right ) }   .
\end{equation}
A change point at time step $t-m$ is identified if the density ratio score $S\left (   \mathbf{\widetilde{r}} _{te} ,\mathbf{\widetilde{r}} _{rf} \right )$ exceeds a predefined threshold $\mu$ (i.e., $S \geq \mu$). Specifically, we adopt the Jensen-Shannon divergence as the density ratio score function, defined as:
\begin{flalign}
\small
\label{eq:14}
S\left (   \mathbf{\widetilde{r}} _{te} ,\mathbf{\widetilde{r}} _{rf} \right ) = & \log 2 + \frac{1}{2n_{te}} \sum_{\mathbf{r}\in \mathbf{\widetilde{r}} _{te} }\log{ P\left ( \mathbf{r}\mid f\left ( \mathbf{r} \right ) =1 \right ) } \notag \\ 
& +\frac{1}{2n_{rf}} \sum_{\mathbf{r}\in \mathbf{\widetilde{r}} _{rf} }\log{ P\left ( \mathbf{r}\mid f\left ( \mathbf{r} \right ) =0 \right ) } .
\end{flalign}
After detecting change points, we propose the \emph{Discrepancy of Environment} (DoE) metric to rigorously quantify the impact of environmental dynamic shifts on state-action value estimates. Specifically, DoE measures the difference in the Q-function for a given state-action pair $(s_k, a_k)$ before and after an environmental change:
\begin{flalign}
\small
\label{eq:10}
DoE\left ( s_k,a_k\right )  & =  \mathbb{E}_{s\sim P_i,a\sim \pi} \left [   { \textstyle \sum_{j=k}^{\infty }}  \gamma^{k-t} R_i\left ( s_j,a_j \right ) \right ] \notag \\ 
& - \mathbb{E}_{s\sim P_{i-1},a\sim \pi} \left [   {\textstyle \sum_{j=k}^{\infty }}  \gamma^{j-t} R_{i-1}\left ( s_j,a_j \right ) \right ] \notag  \\
 & =  Q_i\left ( s_k ,a_k \right ) - Q_{i-1}\left ( s_k,a_k \right )  ,
\end{flalign}
where $Q_{i-1}$ denotes the Q-function from the previous environment $\langle P_{i-1}, R_{i-1} \rangle$, while $Q_i$ represents the current Q-function under the new dynamics $\langle P_i, R_i \rangle$. 

\begin{figure}[t]
  \centering
    \includegraphics[width=0.93\linewidth]{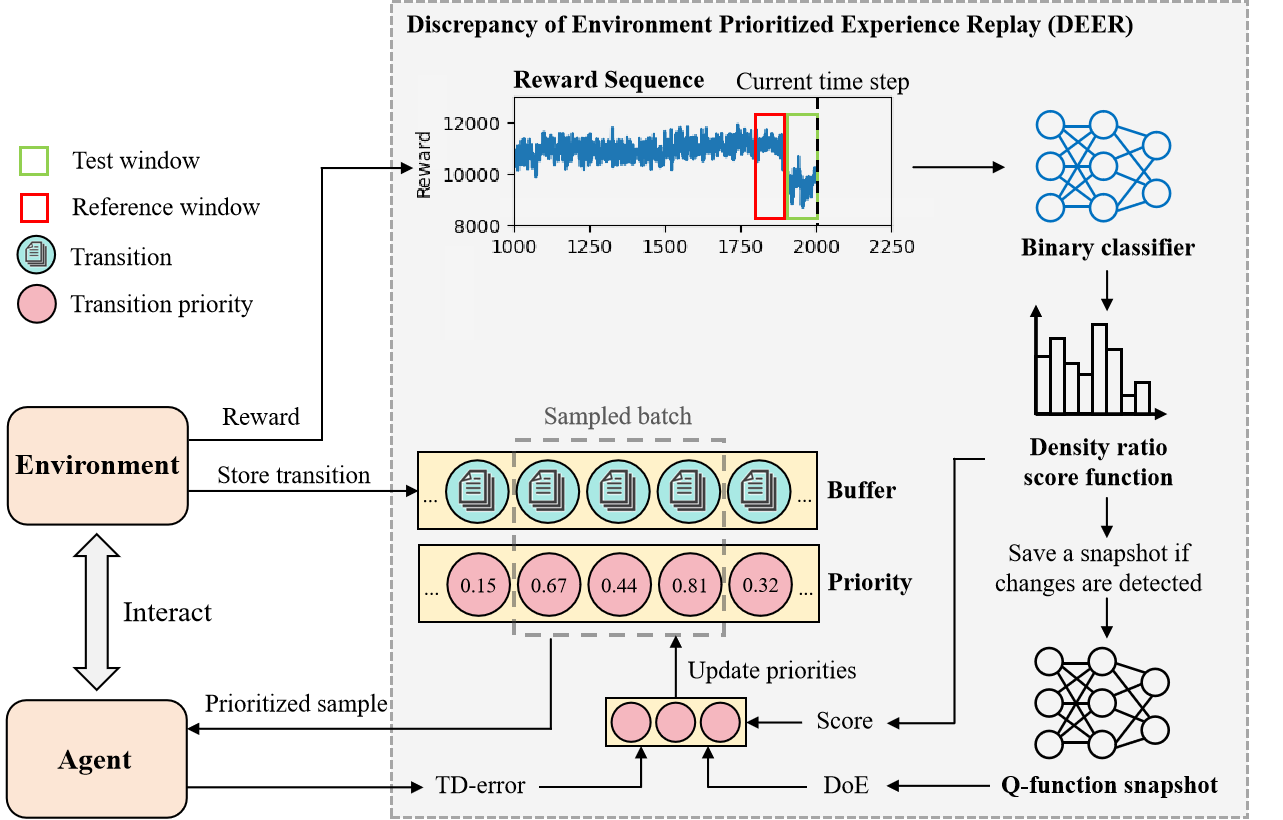}
      \vspace{-1.5ex}
    \caption{The workflow of DEER.}
    \vspace{-1ex}
    \label{fig:Overview.}
\end{figure}

\subsection{Discrepancy of Environment Prioritized Experience Replay (DEER)}
\label{sec:DEER}
We further propose DEER, as illustrated in Figure~\ref{fig:Overview.}. DEER prioritizes pre-change transitions exhibiting low Degree of Environmental change (DoE), as these transitions are less affected by environmental shifts and thus retain their relevance post-change. Specifically, the priority assigned to a transition collected at time step $k$ ($k \leq T_{i-1}$) is defined as:
\begin{equation}
\small
\label{eq:15}
p_k = 2\sigma \left ( -\left | DoE\left ( s_k,a_k\right ) \right |  \right ) ,
\end{equation}
where $p_k$ denotes the priority of the transition collected at time step $k\left ( k\le T_{i-1} \right ) $ and $\sigma$ represents the sigmoid function used to normalize the priority. For post-change transitions, DEER employs a hybrid prioritized sampling strategy guided by real-time density ratio scores that integrate both TD-error and DoE. The density ratio score (Eq.~\ref{eq:14}) quantifies variations in the reward sequence, where a high score indicates ongoing agent adaptation to the altered environment. In such scenarios, transitions with elevated DoE are prioritized to expedite adaptation. Conversely, when the density ratio score is low, sampling emphasizes transitions with higher TD-error to facilitate policy refinement. This adaptive prioritized sampling mechanism is formally defined as follows:
\begin{flalign}
\small
\label{eq:16}
p_{k} = & \left ( 1 -  S\left (   \mathbf{\widetilde{r}} _{te} ,\mathbf{\widetilde{r}} _{rf} \right )\right )  \left ( 2\sigma\left ( \left | TD\left ( s_{k},a_{k},r_{k},s_{k+1} \right )  \right |  \right )  -1\right )  \notag \\
& + S\left (   \mathbf{\widetilde{r}} _{te} ,\mathbf{\widetilde{r}} _{rf} \right ) \left ( 2\sigma\left (  \left | DoE\left ( s_{k},a_{k}\right ) \right |   \right ) -1 \right ) ,
\end{flalign}
where $TD$ denotes the TD-error \cite{schaul2015prioritized}, $p_{k}$ denotes the priority of the transition collected at time step $k\left ( k>  T_{i-1} \right ) $ and $ S\left (   \mathbf{\widetilde{r}} _{te} ,\mathbf{\widetilde{r}} _{rf} \right ) $ denotes the density ratio score at the current time step $t$. To mitigate overfitting caused by frequent replays of high-priority transitions, we normalize the sampling probabilities for both pre- and post-transitions as follows:
\begin{equation}
\small
\label{eq:17}
P\left ( k \right ) = \frac{p^\alpha _k}{ {\textstyle \sum_{i}p^\alpha _i} } ,
\end{equation}
where $\alpha \in (0,1]$ controls the degree of prioritization. Since prioritized sampling changes the distribution used for Q-function estimation, we apply importance sampling to correct the resulting bias. Specifically, the update weight for each transition is computed as:
\begin{equation}
\small
\label{eq:18}
w_k=\frac{1}{\left ( N \cdot P\left ( k \right )  \right ) ^{\beta }\cdot \max_{i} w_i} ,
\end{equation}
where $N$ denotes replay buffer size and $\beta  \in \left ( 0,1 \right ]  $ controls the extent of bias correction.  

\section{Evaluation}
\begin{figure*}[t]
\centering

\subfloat[Ant\label{fig:subfiga}]{
  \includegraphics[width=0.23\textwidth]{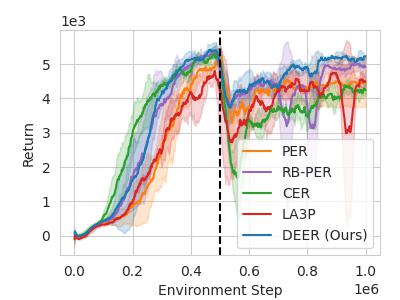}
}
\hfill
\subfloat[HalfCheetah\label{fig:subfigb}]{
  \includegraphics[width=0.23\textwidth]{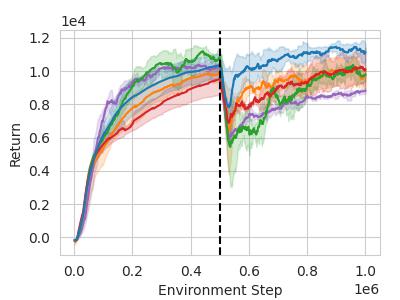}
}
\hfill
\subfloat[Hopper\label{fig:subfigc}]{
  \includegraphics[width=0.23\textwidth]{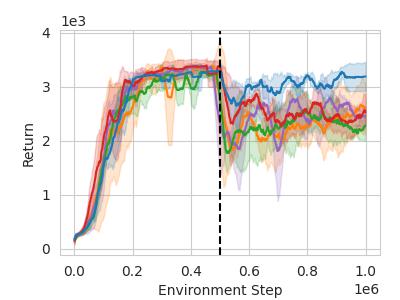}
}
\hfill
\subfloat[ID-Pendulum\label{fig:subfigd}]{
  \includegraphics[width=0.23\textwidth]{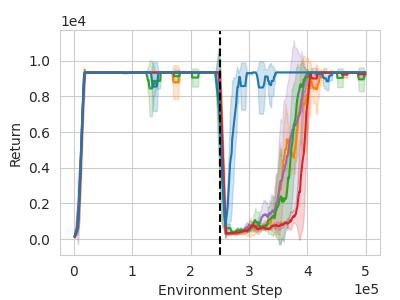}
}

\caption{Sample efficiency comparison of SAC across four tasks. The dashed line marks when the environment dynamics change.}
\label{fig:mainfig}
\end{figure*}
\renewcommand{\arraystretch}{1.1}


\begin{table}[t]
\caption{Average episode rewards over $10^5$ steps post-change for SAC+HalfCheetah under different levels of non-stationarity.}
\label{ns}
\centering
\footnotesize
\begin{tabular}{lllllll}
\hline
Offset & 0\%                        & 50\%                                            & 200\%                     \\ \hline
DEER   & 11894.85 ± 182.36           & \textbf{11789.66 ± 283.81}  & \textbf{9856.27 ± 477.04} \\
PER    & 11834.68 ± 244.31          & 10266.77 ± 327.79                    & 5204.22 ± 763.55          \\
RB-PER & 11754.29 ± 202.61          & 10506.37 ± 364.37                   & 7332.36 ± 318.96          \\
CER    & \textbf{11975.13 ± 256.47} & 9428.88 ± 199.17                      & 8043.76 ± 467.35          \\
LA3P   & 11895.44 ± 267.72          & 11024.15 ± 787.15                    & 6614.79 ± 285.27    \\ \hline
\end{tabular}
\end{table}

\begin{figure}[t]
\centering

\begin{minipage}{0.23\textwidth}
  \centering
  \includegraphics[width=\textwidth]{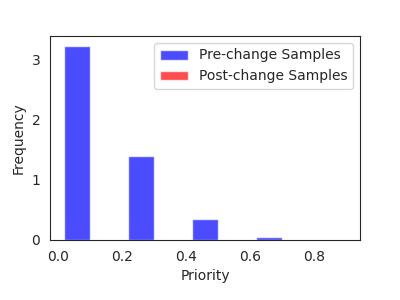}
  \caption*{(a) pre-change $10^4$ steps}
\end{minipage}
\hfill
\begin{minipage}{0.23\textwidth}
  \centering
  \includegraphics[width=\textwidth]{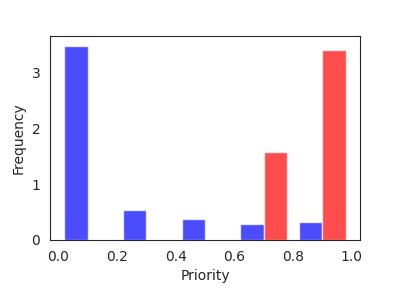}
  \caption*{(b) post-change $10^4$ steps}
\end{minipage}
\vspace{0.4em}

\begin{minipage}{0.23\textwidth}
  \centering
  \includegraphics[width=\textwidth]{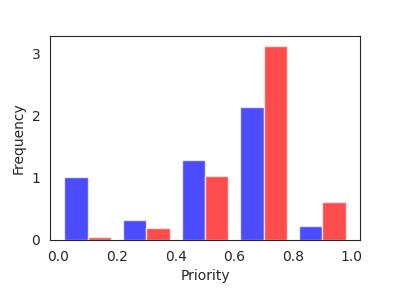}
  \caption*{(c) post-change $10^5$ steps}
\end{minipage}
\hfill
\begin{minipage}{0.23\textwidth}
  \centering
  \includegraphics[width=\textwidth]{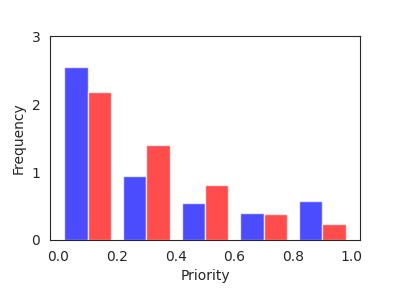}
  \caption*{(d) post-change $5 \times 10^5$ steps}
\end{minipage}

\vspace{-0.4em}
\caption{Priority distribution of pre- and post-change samples at different training stages in SAC+DEER in HalfCheetah.}
\label{fig:distribution}
\vspace{-1em}
\end{figure}

\subsection{Experimental Setup}


We evaluate our proposed integration with the SAC algorithm \cite{haarnoja2018soft} on four standard MuJoCo Gymnasium continuous control tasks: Ant-v4, HalfCheetah-v4, Hopper-v4, and Inverted Double Pendulum-v4 (ID-Pendulum-v4). To simulate non-stationary environments, we introduce offsets to friction and joint damping coefficients—set to the largest values that still permit SAC with uniform sampling to converge. We compare several experience replay (ER) methods, including PER \cite{schaul2015prioritized}, RB-PER \cite{li2021revisiting}, CER \cite{zhang2017deeper}, LA3P \cite{saglam2023actor}, and our proposed DEER. Training is run for $5 \times 10^{5}$ steps on ID-Pendulum and $1 \times 10^{6}$ steps on the other tasks, with non-stationarity introduced halfway. Results are averaged over five runs, with standard deviations as shaded regions. All networks use two hidden layers of 256 units, a learning rate of $1 \times 10^{-3}$, a discount factor of 0.99, a replay buffer size of $1 \times 10^{6}$, and a batch size of 256. For DEER, we set $\alpha=0.6$, $\beta=0.4$, and use a binary classifier (MLP with two 100-unit hidden layers) trained for up to 50 iterations. The detection window is 500, with 10 samples per window (each containing 50 data points), and the detection threshold is 0.5.

\subsection{Experimental Results and Analysis}

Fig.~\ref{fig:subfiga}--\ref{fig:subfigd} show the episode return curves when various ER methods are integrated with SAC algorithm. Notably, in both scenarios, DEER (indicated by the red line) achieves higher overall returns compared to the baselines. Throughout various environmental changes, DEER exhibits less reduction in rewards and faster recovery rates than other methods, indicating its efficiency in adapting to dynamic environments.

To further analyze the sampling behavior of DEER, fig.~\ref{fig:distribution} illustrates the priority distribution of samples in the replay buffer at different training stages. Overall, the priority of post-change transitions is higher than that of pre-change ones, with this disparity being particularly pronounced during the early stages of environmental dynamics changes. With density ratio scores decrease as training progresses, The overall sample distribution tends to revert to pre-change state. This reflects DEER's strategy of balancing between pre- and post-change experience: leveraging new experience early on for rapid policy adjustment, followed by a gradual and selective reuse of old experiences to maintain stability and long-term memory.

To analyze the effects of degrees of non-stationarity, table~\ref{ns} presents the average rewards over 100 episodes following an environmental change in the SAC+HalfCheetah task, evaluated across varying levels of non-stationarity: extreme (200\% offset), mild (50\% offset), and stationary (0\% offset). It is worth noting that standard (100\% offset) corresponds to the result shown in Fig.~\ref{fig:subfigb}. The results indicate that DEER significantly outperforms other methods in highly non-stationary settings, achieving roughly 22.53\% higher rewards than the best baseline at the 200\% offset. This highlights DEER's strong adaptability to drastic environmental changes. Under mild non-stationarity, DEER retains a slight performance edge, while in stationary environments, all approaches perform comparably, suggesting that DEER does not introduce negative effects when the environment remains unchanged.

\section{Conclusion}
We propose DEER, a hybrid metric prioritization method that dynamically adjusts sampling weights to address sample efficiency in non-stationary environments. By prioritizing valuable samples according to current environmental changes, DEER improves sample efficiency. As a potential future direction, we are looking forward to extending our method to improve the performance of various applications such as large language models~\cite{lin2025hsplitlora,fang2024automated,lin2024splitlora}, multi-modal training~\cite{fang2025dynamic,tang2024merit}, distributed machine learning~\cite{zhang2024fedac,lin2025leo,hu2024accelerating,zhang2025lcfed,lin2025sl,lin2024adaptsfl,lyu2023optimal,lin2024efficient}, and autonomous driving~\cite{lin2022tracking,fang2024ic3m,lin2022channel}. 

\bibliographystyle{IEEEbib}
\bibliography{refs}

\end{document}